# A Defect in Dempster-Shafer Theory

Pei Wang
Center for Research on Concepts and Cognition
Indiana University
510 North Fess Street, Bloomington, IN 47408
pwang@cogsci.indiana.edu

## Abstract

By analyzing the relationships among *chance*, *weight of evidence* and *degree of belief*, we show that the assertion "probability functions are special cases of belief functions" and the assertion "Dempster's rule can be used to combine belief functions based on distinct bodies of evidence" together lead to an inconsistency in Dempster-Shafer theory. To solve this problem, we must reject some fundamental postulates of the theory. We introduce a new approach for uncertainty management that shares many intuitive ideas with D-S theory, while avoiding this problem.

## 1 INTRODUCTION

Evidence theory, or Dempster-Shafer (D-S) theory, was developed as an attempt to generalize probability theory by introducing a rule for combining distinct bodies of evidence [Dempster 1967, Shafer 1976].

As a formal system, D-S theory is distinguished from other uncertainty management approaches by [Dempster 1967, Shafer 1976]:

1. A *basic probability assignment*, $m(x)$, defined on the space of the *subsets* of competing hypotheses, rather than directly on the hypotheses themselves. $m(x)$ in turn defines the *degree of belief*, $Bel(A)$, and the *degree of plausibility*, $Pl(A)$, of a set of hypotheses $A$. $Bel(x)$ is a generalization of a probability function $q(x)$.

2. *Dempster's rule of combination*, applied to calculate $m_1 \oplus m_2(x)$ from $m_1(x)$ and $m_2(x)$, where $m_1(x)$ and $m_2(x)$ are based on evidence from distinct sources, and $m_1 \oplus m_2(x)$ is based on the pooled evidence.

In this paper, we argue that there is an inconsistency among the fundamental postulates of D-S theory. Though there are several possible solutions of this problem within the framework of D-S theory, each of them has serious disadvantages. Finally, we briefly introduce a new approach that achieves the goals of D-S theory, yet is still natural and consistent.

## 2 A SIMPLIFIED SITUATION

To simplify our discussion, we address only the simplest non-trivial *frame of discernment* $\Theta = \{H, H'\}$ ($|\Theta| = 1$ is trivial). Since $\Theta$ is exhaustive and exclusive by definition, we have $H' = \bar{H}$ (the negation of $H$). In such a situation, the basic probability assignment $m : 2^\Theta \to [0,1]$ is constrained by

$$m(\emptyset) = 0, \quad m(\{H\}) + m(\{\bar{H}\}) + m(\Theta) = 1.$$

Like $m$, the definitions of $Bel$ and $Pl$ are also simplified when $|\Theta| = 2$. With this simplification, we have

$$\begin{array}{ll} Bel(\emptyset) = 0, & Pl(\emptyset) = 0, \\ Bel(\{H\}) = m(\{H\}), & Pl(\{H\}) = 1 - m(\{\bar{H}\}), \\ Bel(\{\bar{H}\}) = m(\{\bar{H}\}), & Pl(\{\bar{H}\}) = 1 - m(\{H\}), \\ Bel(\Theta) = 1, & Pl(\Theta) = 1. \end{array}$$

As a result, all the information in these functions can be represented by an (ordered) pair

$$< Bel(\{H\}), Pl(\{H\}) >$$

which indicates the relationship between the hypothesis $H$ and the available evidence.

Dempster's rule in this case becomes

$$\begin{aligned} m_1 \oplus m_2(\{H\}) &= \lambda[m_1(\{H\})m_2(\{H\}) \\ &\quad + m_1(\{H\})m_2(\Theta) \\ &\quad + m_1(\Theta)m_2(\{H\})] \\ m_1 \oplus m_2(\{\bar{H}\}) &= \lambda[m_1(\{\bar{H}\})m_2(\{\bar{H}\}) \\ &\quad + m_1(\{\bar{H}\})m_2(\Theta) \\ &\quad + m_1(\Theta)m_2(\{\bar{H}\})] \\ m_1 \oplus m_2(\Theta) &= \lambda[m_1(\Theta)m_2(\Theta)] \qquad (1) \end{aligned}$$

where
$\lambda = [1 - m_1(\{H\})m_2(\{\bar{H}\}) - m_1(\{\bar{H}\})m_2(\{H\})]^{-1}$.

When $m(\{H\})$ or $m(\{\bar{H}\})$ is equal to 0, there is a special case of belief functions: a *simple support function*, in which the evidence points precisely and unambiguously to a single non-empty subset of $\Theta$ [Shafer 1976, page 75]. In this situation, the *degree of support* for $\{H\}$ (or $\{\bar{H}\}$), $s$, is completely determined by the *weight*, $w$, of the evidence.

What is *weight of evidence*? Though Shafer does not give a general method for practically evaluating it, he attaches the following properties to it [Shafer 1976, pages 7, 88]:

1. $w$ is a measurement defined on bodies of evidence, and it takes values on $[0, \infty]$.
2. When two entirely distinct bodies of evidence are combined, the weight of the pooled evidence is the *sum* of the original ones.

When the simple support functions to be combined each provide positive evidence for $H$, we have

$$m_1(\{H\}) = s_1, \quad m_1(\Theta) = 1 - s_1,$$
$$m_2(\{H\}) = s_2, \quad m_1(\Theta) = 1 - s_2.$$

Then Dempster's rule gives

$$\begin{aligned} m_1 \oplus m_2(\{H\}) &= 1 - (1-s_1)(1-s_2) \\ m_1 \oplus m_2(\{\bar{H}\}) &= 0 \\ m_1 \oplus m_2(\Theta) &= (1-s_1)(1-s_2), \end{aligned}$$

which is the same as *Bernoulli's rule of combination* [Shafer 1976, page 76].

Following the postulates that weights of evidence combine additively and that Dempster's rule (and its special case, Bernoulli's rule) is the correct way to combine them, the function $g$, which maps weight of evidence to degree of support, is determined by

$$g(w_1 + w_2) = 1 - (1 - g(w_1))(1 - g(w_2)).$$

The result is [Shafer 1976, page 78]:

$$g(w) = 1 - e^{-w}. \tag{2}$$

This relationship between *weight of evidence* and *degree of support* can be extended to the situation where the two pieces of evidence to be combined are in conflict. That is, one of them supports $H$, whereas the other supports $\bar{H}$. Concretely, they are

$$m_1(\{H\}) = s_1, \quad m_1(\Theta) = 1 - s_1,$$
$$m_2(\{\bar{H}\}) = s_2, \quad m_1(\Theta) = 1 - s_2.$$

Then Dempster's rule gives

$$\begin{aligned} m_1 \oplus m_2(\{H\}) &= \frac{s_1(1-s_2)}{1-s_1 s_2} \\ m_1 \oplus m_2(\{\bar{H}\}) &= \frac{s_2(1-s_1)}{1-s_1 s_2} \\ m_1 \oplus m_2(\Theta) &= \frac{(1-s_1)(1-s_2)}{1-s_1 s_2}. \end{aligned}$$

If we rewrite the result of the combination of conflicting evidence in terms of weights of evidence, and use $w^+$ and $w^-$ to indicate the weight of *positive* and *negative* evidence for $H$, respectively (their sum is the total weight of evidence, $w$), we get a generalization of (2) [Shafer 1976, page 84]:

$$\begin{aligned} Bel(\{H\}) &= \frac{e^{w^+} - 1}{e^{w^+} + e^{w^-} - 1} \\ Pl(\{H\}) &= \frac{e^{w^+}}{e^{w^+} + e^{w^-} - 1} \end{aligned} \tag{3}$$

On the other hand, we can determine $w^+$ and $w^-$ from $Bel(\{H\})$ and $Pl(\{H\})$ when $Bel(\{H\}) < Pl(\{H\})$, and get [Shafer 1976, page 84]

$$\begin{aligned} w^+ &= \log \frac{Pl(\{H\})}{Pl(\{H\}) - Bel(\{H\})} \\ w^- &= \log \frac{1 - Bel(\{H\})}{Pl(\{H\}) - Bel(\{H\})} \end{aligned} \tag{4}$$

The concept of "evidence combination" can be formulated as

$$w^+ = w_1^+ + w_2^+ \quad \text{and} \quad w^- = w_1^- + w_2^- \tag{5}$$

where $w_1^+$ and $w_2^+$, as well as $w_1^-$ and $w_2^-$, come from distinct sources, and $w^+$ ($w^-$) is the weight of the pooled positive (negative) evidence (for $H$).

Given (3), (4), and (5), we can derive the combination rule in terms of $Bel(\{H\})$ and $Pl(\{H\})$. To simplify the formula, let $b$ stand for $Bel(\{H\})$, and $p$ for $Pl(\{H\})$. Thus,

$$\begin{aligned} b &= \frac{b_1 p_2 + b_2 p_1 - b_1 b_2}{1 - b_1(1-p_2) - b_2(1-p_1)} \\ p &= \frac{p_1 p_2}{1 - b_1(1-p_2) - b_2(1-p_1)}, \end{aligned} \tag{6}$$

which is exactly Dempster's rule when $|\Theta| = 2$.

The above inferences show that when "evidence combination" is understood as (5), the combination rule (in the forms of (1) or (6)) and the functions that relate belief functions to weights of evidence (in the forms of (3) or (4)) are mutually determined.

Generally, we have $Bel(\{H\}) + Bel(\{\bar{H}\}) \leq 1$. When $m(\Theta)$ is 0, or $Bel(\{H\}) = Pl(\{H\})$, it is a special case, where $Bel(\{H\}) + Bel(\{\bar{H}\}) = 1$. From (3), it is clear that this happens if and only if $w$ goes to infinite. In [Dempster 1967], Dempster calls such a belief function "sharp," and treats it as "an ordinary probability measure." In [Shafer 1976], Shafer calls it "Bayesian," and writes it as $Bel_\infty(\{H\})$. Shafer also refers to $Bel_\infty(\{H\})$ as the *chance*, or *aleatory probability*, of $H$ [Shafer 1976, pages 16, 33, 201]. [1]

## 3 A PROBLEM

From the above descriptions, D-S theory seems to be a reasonable extension of probability theory because it introduces a combination rule, and still converges to probability theory when $Bel(\{H\})$ and $Pl(\{H\})$ overlap.

To see clearly how D-S theory and probability theory are related to each other, consider the situation where evidence for $H$ is in the form of a sequence of experiment outcomes with the following properties:

1. No single outcome can completely confirm or refute $H$.

---
[1] In this paper, *probability* is always used to indicate what Shafer calls *chance* or *aleatory probability*.

2. There are only two possible outcomes: one supports $H$, while the other supports $\bar{H}$.
3. The outcomes are independent and provide distinct bodies of evidence.

In the following, $t$ represents the number of available outcomes, and $t^+$ is the number of outcomes that support $H$. Obviously, $t \geq t^+ \geq 0$.

There are four fundamental assertions in D-S theory that are accepted as postulates:

**Assertion 1.** $q(H) = \lim_{t \to \infty} \frac{t^+}{t}$: the chance of $H$ is the limit of the proportion of positive outcomes among all outcomes [Shafer 1976, pages 9, 202].

**Assertion 2.** $Bel_\infty(\{H\}) = q(H)$: the chance of $H$ is adopted as degree of belief [Shafer 1976, pages 16, 201].

**Assertion 3.** $w^+ = w_1^+ + w_2^+$ and $w = w_1 + w_2$: evidence combination corresponds to the addition of the weights of evidence from distinct sources [Shafer 1976, pages 8, 77].

**Assertion 4.** Dempster's rule is the correct rule for evidence combination [Shafer 1976, pages 6, 57].

Though the four assertions are reasonable when taken individually, *they are inconsistent collectively*. To show this, let us first study the relationship between *chance* and *weight of evidence*.

Because there are only two types of evidence, we can assign two positive real numbers $w_0^+$ and $w_0^-$ as weights of evidence to an outcome supporting $H$ and $\bar{H}$, respectively. After $t$ outcomes are observed, the weight of available positive, negative and total evidence (for $H$) can be calculated according to Assertion 3:

$$\begin{aligned} w^+ &= w_0^+ t^+, \\ w^- &= w_0^- (t - t^+), \\ w &= w^+ + w^-. \end{aligned}$$

When $t$ goes to infinity so does $w$, and vice versa. If $\frac{t^+}{t}$ converges to a limit $q$, then according to Assertion 1 and Assertion 2, $Bel(\{H\})$ and $Pl(\{H\})$ should also converge to $q$, to become $Bel_\infty(\{H\})$.

We can rewrite $w^+$ and $w^-$ as functions of $t$ and $t^+$ in the relationships between belief function and weight of evidence (3), which is derived from Assertion 3 and Assertion 4. If we then take the limit of the equation when $t$ (as well as $w$) goes to infinity, we get

$$\begin{aligned} Bel_\infty(\{H\}) &= \lim_{w \to \infty} \frac{e^{w^+} - 1}{e^{w^+} + e^{w^-} - 1} \\ &= \begin{cases} 0 & \text{if } w_0^+ q < w_0^- (1 - q) \\ 0.5 & \text{if } w_0^+ q = w_0^- (1 - q) \\ 1 & \text{if } w_0^+ q > w_0^- (1 - q). \end{cases} \end{aligned}$$

This means that if $q$ (the chance of $H$ defined by Assertion 1) exists, then, by repeatedly applying Dempster's rule to combine the coming evidence, both $Bel(\{H\})$ and $Pl(\{H\})$ will converge to a point. However, that point is not $q$ in most cases, but 0, 0.5, or 1, indicating qualitatively whether there is more positive evidence than negative evidence. This result contradicts Assertion 2.

What does a Bayesian belief function correspond to when $Bel_\infty(\{H\})$ is not in $\{0, 0.5, 1\}$? Shafer makes it clear that this happens when $w^- - w^+$ has a constant limit $\Delta$ [Shafer 1976, page 197]. In that case,

$$\begin{aligned} Bel_\infty(\{H\}) &= \lim_{w \to \infty} \frac{e^{w^+} - 1}{e^{w^+} + e^{w^-} - 1} \\ &= \frac{1}{1 + e^\Delta}. \end{aligned}$$

In the current example, this happens when

$$\lim_{t \to \infty} [w_0^-(t - t^+) - w_0^+ t^+] = \Delta.$$

Therefore, though a Bayesian belief function is indeed a probability function in the sense that

$$Bel(\{H\}) + Bel(\{\bar{H}\}) = 1,$$

it is usually different from the chance of $H$. $Bel_\infty(\{H\})$ and $q(H)$ are equal only when

1. $q(H)$ is 0 or 1, or
2. $q(H)$ is 0.5, and $w_0^+ = w_0^-$.

This inconsistency is derived from the four assertions alone, so it is independent from other controversial issues about D-S theory: such as the interpretation of belief function; the accurate definition of "distinct" or "independent" bodies of evidence; and the actual measurement of weight of evidence. No matter what opinions are accepted on these issues, as long as they are held consistently, the previous problem remains. For example, the choice of $w_0^+$ and $w_0^-$ can only determine which chance value is mapped to the degree of belief 0.5 (so all the other values are mapped to 0 or 1 correspondingly), but cannot change the result that chance and Bayesian belief function are usually different.

This discrepancy also unearths some other inconsistencies in D-S theory. For example, Shafer describes *chance* as "essentially hypothetical rather than empirical," and unreachable by collecting (finite) evidence [Shafer 1976, page 202]. According to this interpretation, combining the evidence of two different Bayesian belief functions becomes invalid or nonsense, because they are chances and therefore not supported by finite empirical evidence. If $Bel_{\infty 1}(\{H\})$ and $Bel_{\infty 2}(\{H\})$ are different, then they are two conflicting conventions, and applying Dempster's rule to them is unjustified. If $Bel_{\infty 1}(\{H\})$ and $Bel_{\infty 2}(\{H\})$ are equal, then they are the same convention made from different considerations. In D-S theory, however, they are combined to get a different Bayesian belief function, unless they happen to be 0, 0.5, or 1. Such a result is counter-intuitive [Wilson 1992] and inconsistent with Shafer's interpretation of chance.

There are already many papers on the justification of Dempster's rule [Kyburg 1987, Pearl 1988, Smets 1990, Dubois and Prade 1991, Voorbraak 1991, Wilson 1993], but few of them addresses the relationships among degree of belief, weight of evidence, and probability. Actually weight of evidence is seldom mentioned in the literature of D-S theory. It is possible to describe and justify D-S theory as a mathematical model without mentioning the relationships. Shafer, in his later papers (for example, [Shafer and Tversky 1985, Shafer 1990]), tends to relate belief functions to *reliability of testimony* and *randomly coded message*, rather than to *weight of evidence*. Even so, the problem is *always there*, because it can be derived from the four fundamental assertions, none of which has been explicitly rejected.

## 4 POSSIBLE SOLUTIONS

To get a consistent theory, at least one of the four assertions must be removed. In the following, let us check all four logical possibilities one by one.

It seems unpopular to reject Assertion 1, and redefine probability as $\lim_{w \to \infty} \frac{e^{w^+} - 1}{e^{w^+} + e^{w^-} - 1}$, though this will lead to a consistent theory. The reason is simple: to use "probability" for the limit of the proportion of positive evidence is a well accepted convention, and a different usage of the concept will cause many confusions.

How about Assertion 3? In the following, we can see that if the *addition* of weight of evidence, during the combination of evidence from distinct sources, is replaced by *multiplication*, we can also get a consistent theory.

Let us assume $w^+ = w_1^+ w_2^+$ and $w^- = w_1^- w_2^-$ when two Bayesian belief functions $Bel_{\infty 1}(\{H\})$ and $Bel_{\infty 2}(\{H\})$ are combined. To simplify the notation, the two functions are written as $b_1$ and $b_2$ in the following. Now, if we simply use the number of outcomes as weight of evidence, then from Assertion 1, Assertion 2, and the new assumption, we get

$$\begin{aligned}
b_1 &= \lim_{w_1 \to \infty} \frac{w_1^+}{w_1}, \\
b_2 &= \lim_{w_2 \to \infty} \frac{w_2^+}{w_2}, \\
b &= \lim_{w \to \infty} \frac{w_1^+ w_2^+}{w_1^+ w_2^+ + w_1^- w_2^-} \\
&= \frac{b_1 b_2}{b_1 b_2 + (1 - b_1)(1 - b_2)}.
\end{aligned}$$

The last equation is a special case of (6), when $b_1 = p_1$ and $b_2 = p_2$ (for Bayesian belief functions).

Though we preserve consistency, the result is not intuitively appealing. For example, no matter how the weight of evidence is actually measured, the combination of two pieces of positive evidence with unit weight ($w_1^+ = w_2^+ = 1$) will get $w^+ = 1$. That is, evidence is no longer accumulated by combination ($w^+$ may even be less than $w_1^+$, if $w_2^+ < 1$). This is not what we have in mind when talking about evidence combination or pooling.

The rejection of Assertion 2 seems more plausible than the previous alternatives. Very few authors actually use $Bel_{\infty}(\{H\})$ to represent the probability of $H$. Even in Shafer's classic book [Shafer 1976], in which Assertion 2 was made or assumed at several places, $Bel_{\infty}(x)$ is not directly applied to represent statistical evidence.

However, there is not a consensus in the "Uncertainty in AI" community that $Bel_{\infty}(x)$ and $q(x)$ are different. The following phenomena shows this:

1. The "lower-upper bounds of probability" interpretation for belief functions is still accepted by some authors [Fagin and Halpern 1991].

2. Some other authors, including Shafer himself, reject the above interpretation, but they still refer to a probability function as a special type (a limit) of a belief function [Shafer 1990].

3. Though some authors have gone so far to the conclusion that Bayesian belief functions do not generally correspond to Bayesian measures of belief, they still view a belief function as the lower bound of probability [Wilson 1992].

Even if everyone agrees that $Bel_{\infty}(x)$ and $q(x)$ are different measurements of uncertainty and that all relationships between probability theory and D-S theory are cancelled, as Smets outlines in the transferable belief model of D-S theory [Smets 1990, Smets 1991], there are still problems with this model.

We have shown that, following Assertion 1, Assertion 3, and Assertion 4, $q(H)$ not only is different from $Bel_{\infty}(\{H\})$, but also cannot be properly represented in a belief function or a basic probability assignment. The proportion of positive evidence of $H$ can be derived from $Bel(\{H\})$ and $Pl(\{H\})$, when $Bel(\{H\}) < Pl(\{H\})$, as

$$\frac{w^+}{w} = \frac{\log p - \log(p - b)}{\log p + \log(1 - b) - 2\log(p - b)}$$

where $b$ is $Bel(\{H\})$ and $p$ is $Pl(\{H\})$. Still, the relationship is not natural, and the ratio usually does not converge to the same point with $Bel(\{H\})$ and $Pl(\{H\})$. As a result, a natural way to represent uncertainty as *proportion of positive evidence* becomes less available in D-S theory. As shown before, $Bel(\{H\})$ is more sensitive to the difference of $w^+$ and $w^-$, than to the proportion $\frac{w^+}{w}$. $q(x)$, as the limit of the proportion, even cannot be represented. The knowledge "$q(H) = 0.51$" and "$q(H) = 0.99$" will both be represented as $Bel(\{H\}) = Pl(\{H\}) = 1$, and their difference will be lost. [2]

If Assertion 2 were rejected, it would be invalid to interpret $Bel(\{H\})$ and $Pl(\{H\})$ as "lower and upper probability" [Dempster 1967, Smets 1991, Fagin and Halpern 1991, Dubois and Prade 1992]. It is true that there are probability functions $P(x)$ satisfying

$$Bel(\{x\}) \leq P(x) \leq Pl(\{x\}), \text{ for all } x \in \Theta.$$

However, as demonstrated above, these functions may be unrelated to $q(H)$.

---

[2] Here we do not distinguish $\frac{w^+}{w}$ and $\frac{t^+}{t}$, because their difference does not influence the conclusion, as shown previously.

For the same reason, the assertion that "the Bayesian theory is a limiting case of D-S theory" [Shafer 1976, page 32] may be misleading. From a mathematical point of view, this assertion is true, since $Bel_\infty(\{H\})$ is a probability function. But as discussed previously, it is not the probability of $H$. Therefore, it is not valid to get inference rules for D-S theory by extending Bayes theorem. In general, the relationship between D-S theory and probability theory will be very loose.

It is still possible to put different possible probability distributions into $\Theta$ and to assign belief function to them, as Shafer did [Shafer 1976, Shafer 1982]. For example, the knowledge "$q(H) = 0.51$" can be represented as "$Bel(\{q(H) = 0.51\}) = 1$." However, here the probability function is *evaluated* by the belief function, rather than being a special case of it. The two are at different levels. As a result, the initial idea of D-S theory (to generalize probability theory), no longer holds. From a practical point of view, this approach is not appealing, neither. For instance, for any evidence combination to occur there must be *finite* possible probabilities for $H$ at the very beginning. It is unclear how to get them.

Finally, it is unlikely, though not completely impossible, to save D-S theory by rejecting Assertion 4. We can say that Dempster's rule does not apply to evidence combination, but can be used for some other purposes. Even so, the initial goal of D-S theory will be missed.

In summary, though it is possible for D-S theory to survive the inconsistency by removing one of the assertions, the result is still unsatisfactory. Either the natural meaning of "probability" or "evidence combination" must be changed, or the theory will fail to meet its original purpose, that is, to extend probability theory by introducing an evidence combination rule.

## 5  AN ALTERNATIVE APPROACH

In spite of the problems, some intuitions behind D-S theory are still attractive, such as the first three assertions, the idea of lower-upper probabilities [Dempster 1967], and the distinction between disbelief and lack of belief [Shafer 1976].

From previous discussion, we have seen that the core of evidence combination is the relationships among degree of belief, probability, and weight of evidence. The combination rule can be derived from the relationships.

Let us continue with the previous example. Because all the measurements are about $H$, we will omit it to simplify the formulas. Following the practice of statistics, for the current example a very natural convention is to use the number of outcomes as the weight of evidence: $w_0^+ = w_0^- = 1$.

Because our belief about $H$ is totally determined by available evidence, it may be uncertain due to the existence of negative evidence. To measure the relative support that $H$ gets from available evidence, the most often used method is to take the *frequency* of positive evidence: $f = \frac{w^+}{w}$. According to Assertion 1, $\lim_{w \to \infty} f = q$, that is, the limit of $f$, if it exists, is the probability of $H$. Therefore, we can refer to frequency as probability generalized to the situation of finite evidence.

However, when evidence combination is considered, $f$ alone cannot capture the uncertainty about $H$. When new evidence is combined with previous evidence, $f$ must be reevaluated. If we only know its previous value, we cannot determine how much it should be changed — the absolute amount of evidence is absent in $f$. Can we capture this kind of information without recording $w$ and $w^+$ directly? [3]

Yes, we can. From the viewpoint of evidence combination, the influence of $w$ appears in the *stability* of a frequence evaluation based on it. Let us compare two situations: in the first, $w = 1000$ and $w^+ = 600$, and in the other $w = 10$ and $w^+ = 6$. Though in both cases $f$ is 0.6, the stability is quite different. After a new outcome is observed, in the first situation the new frequency becomes either $\frac{600}{1001}$ or $\frac{601}{1001}$, while in the second it is $\frac{6}{11}$ or $\frac{7}{11}$. The adjustment is much larger in the second situation than in the first.

If the information about stability is necessary for evidence combination, why not directly use intervals like $[\frac{600}{1001}, \frac{601}{1001}]$ and $[\frac{6}{11}, \frac{7}{11}]$ to represent the uncertainty in the previous situations?

Generally, let us introduce a pair of new measurements: a *lower frequency*, $l$, and a *upper frequency*, $u$, which are defined as

$$\begin{aligned} l &= \frac{w^+}{w+1} \\ u &= \frac{w^+ + 1}{w+1}. \end{aligned} \quad (7)$$

The idea behind $l$ and $u$ is simple: if the current frequency is $\frac{w^+}{w}$, then, after combining the current evidence (whose weight is $w$) with the new evidence provided by a new outcome (whose weight is 1), the new frequency will be in the interval $[l, u]$. [4]

As bounds of frequency, $l$ and $u$ share intuitions with Dempster's $P_*$ and $P^*$, as well as Shafer's $Bel$ and $Pl$. However, they have some properties that distance them from the functions of D-S theory and other similar ideas like lower and upper bounds of probability:

1. $l \leq f \leq u$, that is, the current frequency is within the $[l, u]$ interval. Furthermore, it is easy to see that $f = \frac{l}{1-u+l}$, so the frequency value can be easily retrieved from the bounds.

---

[3]Though it is possible, in theory, to directly use $w$ and $w^+$ as measurements of uncertainty, it is often unnatural and inconvenient. See [Wang 1993b] for more discussions.

[4]We use an interval instead of a pair of points because the measurements will be extended to situations in which the weights of evidence are not necessarily integers. In general, the interval bounds the frequence until the weight of new evidence reaches a constant unit. For the current purpose, the 1 that appears in the definitions of $l$ and $u$ can be substituted by any positive number. 1 is used here to simplify the discussion. See [Wang 1993b].

2. The bounds of frequency are defined in terms of available evidence, which is finite. Whether the frequence of positive evidence really has a limit does not matter. On the other hand, the interval can be determined before the next outcome occurs.

3. $\lim_{w \to \infty} l = \lim_{w \to \infty} f = \lim_{w \to \infty} u = q$. If $f$ does have a limit $q$, then $q$ is also the limit of $l$ and $u$. Therefore, probability is a special case of the $[l, u]$ interval, in which the interval degenerates into a point.

4. However, $q$, if it exists, is not necessarily in the interval all the time that evidence is accumulating. $[l, u]$ indicates the range $f$ will be from the current time to a *near future* (until the weight of new evidence reaches a constant), not an *infinite future*. Therefore, $l$ and $u$ are not bounds of probability.

5. The width of the interval $i = u - l = \frac{1}{w+1}$ monotonically decreases during the accumulating of evidence, and so can be used to represent the system's "degree of ignorance" (about $f$). When $w = 0, i = 1$, because with no evidence, ignorance reaches its maximum. When $w \to \infty, i = 0$, because with infinite evidence the probability is obtained, so the ignorance (about the frequency) reaches its minimum, even though the next outcome is still uncertain. In this way, "lack of belief" and "disbelief" are clearly distinguished.

From the definitions of the lower-upper frequencies and Assertion 3, a combination rule, from $[l_1, u_1] \times [l_2, u_2]$ to $[l, u]$, is uniquely determined in terms of lower-upper frequencies, when neither $i_1 = u_1 - l_1$ nor $i_2 = u_2 - l_2$ is 0:

$$l = \frac{l_1 i_2 + l_2 i_1}{i_1 + i_2 - i_1 i_2}$$
$$u = \frac{l_1 i_2 + l_2 i_1 + i_1 i_2}{i_1 + i_2 - i_1 i_2}. \quad (8)$$

From (3) and (7), we can even set up a one-to-one mapping between the *Bel-Pl* scale and the $l$-$u$ scale, when the weight of evidence $w$ is finite and $|\Theta| = 2$. In this way, the combination rule given by (8) is mapped exactly onto Dempster's rule (6). From a mathematical point of view, the two approaches differ only when $w \to \infty$. Then *Bel* and *Pl* converge to a non-trivial (not in $\{0, 0.5, 1\}$) probability if and only if $w^- - w^+$ converges to a constant, but $l$ and $u$ converge to a non-trivial probability if and only if $\frac{w^+}{w}$ converges to a constant. The latter, being the probability of $H$, is more helpful and important in most situations than the former is. In fact, Shafer acknowledges the problem when he writes, "It is difficult to imagine a belief function such as $Bel_\infty$ being useful for the representation of actual evidence [Shafer 1976, page 199]." However, the result seems to be accepted without further analysis, since it follows from Dempster's rule.

Let us apply the paradigm to infinite evidence. For practical purpose it is impossible for a system to get infinite evidence, but we can use this concept to put definitions and conventions into a system. Beliefs supported by infinite evidence can be processed as normal ones, but will not be changed through evidence combinations.

According to the interpretation of the $[l, u]$ interval, it is not difficult to extend the new combination rule (8) to the case of infinite evidence:

1. When $i_1 = 0$ but $i_2 > 0$, the rule is still applicable in the form of (8), which gives the result that $l = l_1 = u_1 = u$. Thus when uncertainty is represented by probability (a point, instead of an interval), it will not be effected by combining its evidence with finite new evidence.

2. When $i_1 = i_2 = 0$, the rule cannot be used. Now the system will distinguish two cases:
   (a) when $l_1 = l_2 = u_1 = u_2$ there are two identical probabilistic judgments, so one of them can be removed (because it is redundant), leaving the other as the conclusion; or,
   (b) $l_1 \neq l_2$, meaning there are two conflicting probabilistic judgments. Since such judgments are not generated from evidence collection but from conventions or definitions, the two judgments are not "combined," but reported to the human/program which is responsible for making the conventions.

Here we are even more faithful to Shafer's interpretation of (aleatory) probability than D-S theory is. Being "essentially hypothetical rather than empirical," probability cannot be evaluated with less than infinite evidence [Shafer 1976, page 201]. For the same reason, it should not be changed by less than infinite evidence.

In summary, though many of the intuitive ideas of D-S theory are preserved, the problem in D-S theory discussed above no longer exists in the "lower-upper frequency" approach. The new method can represent probability and ignorance, and has a rule for evidence combination. The new approach can hardly be referred to as a modification or extension of D-S theory, in part because Dempster's rule is not used.

This approach is used in the *Non-Axiomatic Reasoning System* (NARS) project. As an intelligent reasoning system, NARS can adapt to its environment and answer questions with insufficient knowledge and resources [Wang 1993a, Wang 1993b]. A complete comparison of NARS and D-S theory is beyond the scope of this paper. By introducing the approach here, I hope to show that the most promising solution for the previous inconsistency is to choose a new rule for evidence combination.

## 6 CONCLUSION

Though the criticism of D-S theory to Bayes approach is justifiable, and the "lower-upper frequency" approach is motivated by similar theoretical considerations [Wang 1993c], the two approaches solve the problem differently.

D-S theory, though it can be used to accumulate evidence from distinct sources, establishes a unnatural relation between degree of belief and weight of evidence by using Dempster's rule for evidence combination. As a result, the assertion that "probability is a special belief function" is in

566  Wang

conflict with the definitions of "probability" and "evidence combination."

The inconsistency is solvable within D-S theory, but such a solution will make D-S theory either lose its naturalness (by using a concept in a unusual way), or miss its original goals (by being unable to represent probability or to combine evidence).

Though not specially designed to replace D-S theory in general, the "lower-upper frequency" approach does suggest a better way to represent and process uncertainty. The new approach sets up a more natural relation among the various measurements of uncertainty, including probability. It can combine evidence from distinct sources. Therefore, it makes the system capable of carrying out multiple types of inference, such as deduction, induction, and abduction [Wang 1993a, Wang 1993b].

### Acknowledgment

This work is supported by a research assistantship from Center for Research on Concepts and Cognition, Indiana University. Thanks to Angela Allen for polishing my English. I also appreciate the helpful comments of the anonymous referees.

## References

[Dempster 1967] A. Dempster. Upper and lower probabilities induced by a multivalued mapping. *Annals of Mathematical Statistics*, 38:325–339, 1967.

[Dubois and Prade 1991] D. Dubois and H. Prade. Updating with belief functions, ordinal conditional functions and possibility measures. In P. Bonissone, M. Henrion, L. Kanal, and J. Lemmer, editors, *Uncertainty in Artificial Intelligence 6*, pages 311–329. North-Holland, Amsterdam, 1991.

[Dubois and Prade 1992] D. Dubois and H. Prade. Evidence, knowledge, and belief functions. *International Journal of Approximate Reasoning*, 6:295–319, 1992.

[Fagin and Halpern 1991] R. Fagin and J. Halpern. A new approach to updating beliefs. In P. Bonissone, M. Henrion, L. Kanal, and J. Lemmer, editors, *Uncertainty in Artificial Intelligence 6*, pages 347–374. North-Holland, Amsterdam, 1991.

[Kyburg 1987] H. Kyburg. Bayesian and non-Bayesian evidential updating. *Artificial Intelligence*, 31:271–293, 1987.

[Pearl 1988] J. Pearl. *Probabilistic Reasoning in Intelligent Systems*. Morgan Kaufmann Publishers, San Mateo, California, 1988.

[Shafer 1976] G. Shafer. *A Mathematical Theory of Evidence*. Princeton University Press, Princeton, New Jersey, 1976.

[Shafer 1982] G. Shafer. Belief functions and parametric models. *Journal of the Royal Statistical Society. Series B*, 44:322–352, 1982.

[Shafer 1990] G. Shafer. Perspectives on the theory and practice of belief functions. *International Journal of Approximate Reasoning*, 4:323–362, 1990.

[Shafer and Tversky 1985] G. Shafer and A. Tversky. Languages and designs for probability judgment. *Cognitive Science*, 12:177–210, 1985.

[Smets 1990] Ph. Smets. The combination of evidence in the transferable belief model. *IEEE Transactions on Pattern Analysis and Machine Intelligence*, 12:447–458, 1990.

[Smets 1991] Ph. Smets. The transferable belief model and other interpretations of Dempster-Shafer's model. In P. Bonissone, M. Henrion, L. Kanal, and J. Lemmer, editors, *Uncertainty in Artificial Intelligence 6*, pages 375–383. North-Holland, Amsterdam, 1991.

[Voorbraak 1991] F. Voorbraak. On the justification of Dempster's rule of combination. *Artificial Intelligence*, 48:171–197, 1991.

[Wang 1993a] P. Wang. Non-axiomatic reasoning system (version 2.2). Technical Report 75, Center for Research on Concepts and Cognition, Indiana University, Bloomington, Indiana, 1993.

[Wang 1993b] P. Wang. From inheritance relation to non-axiomatic logic. Technical Report 84, Center for Research on Concepts and Cognition, Indiana University, Bloomington, Indiana, 1993.

[Wang 1993c] P. Wang. Belief revision in probability theory. In *Proceedings of the Ninth Conference on Uncertainty in Artificial Intelligence*, pages 519–526. Morgan Kaufmann Publishers, San Mateo, California, 1993.

[Wilson 1992] N. Wilson. The combination of belief: when and how fast? *International Journal of Approximate Reasoning*, 6:377–388, 1992.

[Wilson 1993] N. Wilson. The assumptions behind Dempster's rule. In *Proceedings of the Ninth Conference on Uncertainty in Artificial Intelligence*, pages 527–534. Morgan Kaufmann Publishers, San Mateo, California, 1993.